\documentclass[11pt]{article}

\usepackage{amsmath,amssymb}
\usepackage{graphicx}
\usepackage{algorithm}
\usepackage{algpseudocode}
\usepackage{booktabs}
\usepackage{subcaption}
\usepackage{multirow}
\usepackage{tabularx}
\usepackage{makecell}
\usepackage{xcolor}
\usepackage{placeins}

\usepackage[colorlinks=true,
            linkcolor=blue,
            citecolor=blue,
            urlcolor=blue]{hyperref}

\usepackage[margin=1in]{geometry}

\title{\bf Meta-Cognitive Reinforcement Learning with Self-Doubt and Recovery}

\author{
Zhipeng Zhang$^{1,2}$,
Xiongfei Su$^{1}$,
Kai Li$^{1}$
}
\date{
\small
$^{1}$ China Mobile Research Institute, Beijing, China\\
$^{2}$ China Mobile GBA (Greater Bay Area) Innovation Institute, Guangzhou, China\\
Corresponding author: \texttt{zhangzhipeng@chinamobile.com}
}

\begin{document}
\maketitle

% ==================== Abstract ====================
\begin{abstract}
Robust reinforcement learning methods typically focus on suppressing unreliable
experiences or corrupted rewards, but they lack the ability to reason about the
reliability of their own learning process. As a result, such methods often either
overreact to noise by becoming overly conservative or fail catastrophically when
uncertainty accumulates.

In this work, we propose a meta-cognitive reinforcement learning framework that
enables an agent to assess, regulate, and recover its learning behavior based on
internally estimated reliability signals. The proposed method introduces a meta-trust
variable driven by Value Prediction Error Stability (VPES), which modulates learning
dynamics via fail-safe regulation and gradual trust recovery.

Experiments on continuous-control benchmarks with reward corruption demonstrate
that recovery-enabled meta-cognitive control achieves higher average returns and
significantly reduces late-stage training failures compared to strong robustness
baselines.
\end{abstract}


% ==================== Main Content ====================
\section{Introduction}\label{sec:introduction}

Reinforcement learning (RL) systems deployed in real-world settings must operate under unreliable feedback, non-stationary environments, and internal training instability. In practice, such systems often exhibit late-stage degradation or catastrophic collapse, leading to wasted computational resources and unreliable deployment outcomes. These challenges are well documented in studies of real-world reinforcement learning, where training sensitivity and irreproducibility remain persistent obstacles \cite{henderson2018deep, dulac2021challenges}. The problem is particularly pronounced in long-horizon and safety-critical learning systems, where failures may be difficult to detect and recover from once they occur.

A substantial body of work on robust reinforcement learning seeks to mitigate these issues by suppressing corrupted rewards, down-weighting unreliable experiences, or regularizing optimization dynamics \cite{everitt2017reinforcement, moos2022robust, al2018continuous}. While such approaches improve robustness at the level of data or gradients, they implicitly treat learning as an always-on process, even under prolonged internal uncertainty. As a result, they lack explicit mechanisms for assessing whether the learning process itself is currently reliable, often leading to either overly conservative behavior or uncontrolled instability when uncertainty accumulates.

From a systems perspective, effective learning under uncertainty requires not only robustness to noise, but also the ability to regulate when and how learning should occur. In many real-world learning systems, continued optimization during periods of internal instability can be more harmful than beneficial. This observation motivates a shift from designing increasingly sophisticated robustness heuristics toward endowing agents with explicit mechanisms to monitor internal learning stability and regulate learning permissibility accordingly. Such regulation resonates with the notion of meta-cognition in cognitive science, which concerns the monitoring and control of one's own cognitive processes \cite{flavell1979metacognition}.

In this work, we propose a \emph{meta-cognitive reinforcement learning framework} that enables an agent to assess, regulate, and recover its learning behavior based on internal stability signals. Here, meta-cognition—and the associated notion of ``self-doubt''—refers to an internally estimated reliability signal derived from training dynamics, rather than a psychological construct. As outlined in Section~\ref{sec:system_overview}, the framework introduces a meta-trust variable that reflects confidence in the current learning dynamics and is driven by Value Prediction Error Stability (VPES). When instability is detected, a fail-safe regulation mechanism suppresses aggressive updates to prevent catastrophic divergence. Crucially, we further incorporate a recovery mechanism that allows meta-trust—and hence learning capacity—to be gradually restored as internal stability improves. Unlike adaptive optimization or learning-rate scheduling methods, the proposed approach regulates the permissibility of learning itself rather than accelerating convergence.

The main contributions of this work are summarized as follows:
\begin{itemize}
\item \textbf{Problem identification:} We identify a fundamental limitation of existing robust reinforcement learning methods: the absence of explicit mechanisms for assessing and regulating trust in the learning process itself, particularly for recovery from instability.
\item \textbf{Framework design:} We propose a meta-cognitive reinforcement learning framework that integrates internal stability monitoring (via VPES), fail-safe regulation, and trust recovery within a unified control loop applicable to existing RL algorithms.
\item \textbf{Empirical validation:} We empirically demonstrate that recovery-enabled meta-cognitive control improves both performance and robustness, substantially reducing late-stage training failures under reward corruption while maintaining competitive sample efficiency.
\end{itemize}

\section{Related Work}\label{sec:related_work}

\subsection{Robust Reinforcement Learning under Uncertainty}
Robust reinforcement learning has been extensively studied to address uncertainty arising from noisy rewards, adversarial perturbations, or non-stationary environments \cite{moos2022robust, everitt2017reinforcement, al2018continuous}. Prior work has proposed techniques such as robust optimization objectives, uncertainty-aware value estimation, reward clipping, and adversarial training to mitigate the impact of corrupted feedback \cite{pinto2017robust, mankowitz2020robust, engstrom2020impl, zhang2020robust}. Other approaches down-weight unreliable samples using variance-based or disagreement-based criteria, improving stability during policy optimization \cite{kumar2020discor, pathak2019self}.

Safe reinforcement learning focuses on avoiding catastrophic behaviors by enforcing constraints, limiting policy updates, or penalizing unsafe actions \cite{garcia2015comprehensive, achiam2017constrained, dalal2018safe}. Conservative learning strategies often prioritize stability over performance, particularly in safety-critical domains \cite{kumar2020conservative, fujimoto2019off, brunke2022safe, gu2024review}. Such methods have been applied to autonomous driving \cite{kiran2021deep} and robotic control \cite{brunke2022safe, kober2013reinforcement}.

While these methods enhance robustness at the level of data or gradients, they typically treat robustness as a static property of the learning algorithm. They do not explicitly reason about whether the \emph{learning process itself} should be trusted at a given time, nor do they provide mechanisms to suspend and later restore learning based on internal stability. Our work differs by elevating robustness to a meta-cognitive level, where the agent regulates its learning dynamics based on internal signals.

\subsection{Learning Instability and Failure Modes}
Training instability and late-stage collapse have been extensively reported in deep reinforcement learning, particularly in real-world and long-horizon settings \cite{henderson2018deep, dulac2021challenges}. These studies document severe sensitivity to initialization, hyperparameters, and environmental perturbations, with failures often manifesting as abrupt performance degradation after prolonged periods of seemingly stable learning. Despite their empirical prevalence, such failure modes are rarely addressed explicitly by existing robust RL methods, which typically focus on mitigating external uncertainty rather than assessing the reliability of the learning process itself.

\subsection{Adaptive Optimization and Stabilization Techniques}
Meta‑learning approaches aim to adapt learning rules, learning rates, or update strategies based on experience. Examples include learned optimizers, adaptive step‑size methods \cite{xu2019learning}, and higher‑order gradient techniques \cite{xu2018meta, park2020meta}. In reinforcement learning, meta‑gradient methods have been used to tune hyperparameters online or adapt to non‑stationarity \cite{zahavy2020self, al2018continuous}.

Estimating uncertainty in reinforcement learning has been explored through Bayesian methods, ensemble techniques, and disagreement-based metrics \cite{ghavamzadeh2015bayesian, gal2016dropout, osband2016deep, lakshminarayanan2017simple, abdar2021review}. Such methods often use uncertainty estimates to guide exploration or to regularize value updates. Temporal-difference error statistics have also been employed as indicators of learning stability or convergence \cite{fedus2020revisiting}.

\textbf{In contrast to these performance‑oriented adaptations}, our framework does not primarily seek to accelerate convergence or improve asymptotic returns. Instead, the meta‑trust state functions as a \emph{cognitive monitor} that governs the permissibility of learning itself. A typical adaptive optimizer might increase the learning rate when gradient variance is low (suggesting reliable updates). Our meta‑cognitive controller, however, might \emph{decrease} the effective learning rate when VPES is high—even if gradients appear clean—because internal value predictions are inconsistent, signaling broader epistemic uncertainty. 
Crucially, this distinction implies that meta-cognitive control cannot be reduced to adaptive learning-rate heuristics, as it governs the permissibility of learning rather than its optimization efficiency.

\subsection{Cognitive and Meta-Cognitive Perspectives}
From a cognitive science perspective, meta-cognition refers to the ability to monitor and regulate one's own cognitive processes \cite{flavell1979metacognition, nelson1990metamemory}. While reinforcement learning has drawn inspiration from cognitive models of decision-making and learning \cite{niv2009reinforcement, lake2017building}, explicit meta-cognitive mechanisms remain rare.

Recent work has begun exploring self-monitoring and introspection in learning agents, but often lacks concrete control mechanisms tied to learning dynamics \cite{rabinowitz2018machine, griffiths2019doing}. Our work contributes to this emerging direction by providing a concrete, implementable meta-cognitive control loop grounded in internal stability signals and validated through empirical evaluation.

\subsection{Summary}
In summary, existing work on robust, uncertain, meta-learned, and safe reinforcement learning addresses important aspects of learning under uncertainty, but does not fully capture the dynamics of self-doubt and recovery. Robust RL methods focus on filtering unreliable data but lack explicit mechanisms for assessing trust in the learning process. Adaptive optimization techniques aim to improve convergence but do not regulate whether learning should proceed. Safe RL prioritizes constraint satisfaction but often enforces permanent conservatism. Our approach bridges these areas by introducing a meta-cognitive framework that regulates learning based on internal confidence and restores trust when evidence supports renewed stability, enabling sustained adaptation in uncertain environments.

\section{Problem Formulation and System Overview}\label{sec:system_overview}

\subsection{Learning Instability as a System-Level Problem}
In long-horizon reinforcement learning deployments, training instability often manifests as late-stage performance collapse, where an agent's policy suddenly degrades after periods of stable learning. This phenomenon is particularly problematic in safety-critical and real-world systems, where recovery mechanisms are limited and failures are costly \cite{dulac2021challenges, henderson2018deep}. We frame this challenge as a system-level control problem: the learning process itself must be regulated to prevent irreversible divergence and enable sustained adaptation under uncertainty.

Traditional robust reinforcement learning methods address specific sources of uncertainty (e.g., noisy rewards or perturbed observations) but do not provide explicit mechanisms for assessing whether the learning dynamics themselves are currently reliable \cite{moos2022robust, everitt2017reinforcement}. As a result, continued optimization during periods of internal instability can lead to catastrophic updates that are difficult to recover from. This observation aligns with studies showing that RL training is highly sensitive to hyperparameters and environmental conditions \cite{henderson2018deep}, and that failures in real-world deployments often stem from unmanaged learning instability rather than algorithmic deficiencies \cite{dulac2021challenges}.

%This motivates a shift from designing increasingly sophisticated robustness heuristics toward endowing agents with explicit mechanisms to monitor internal learning stability and regulate learning permissibility accordingly. 
Such regulation resonates with control-theoretic approaches to system stability \cite{khalil2002nonlinear} and finds practical importance in applications like autonomous driving \cite{kiran2021deep} and robotic control \cite{brunke2022safe}, where learning must be carefully managed to ensure operational safety.

\subsection{System Architecture and Control Loop}
\begin{figure}[htbp]
    \centering
    \includegraphics[width=0.95\linewidth]{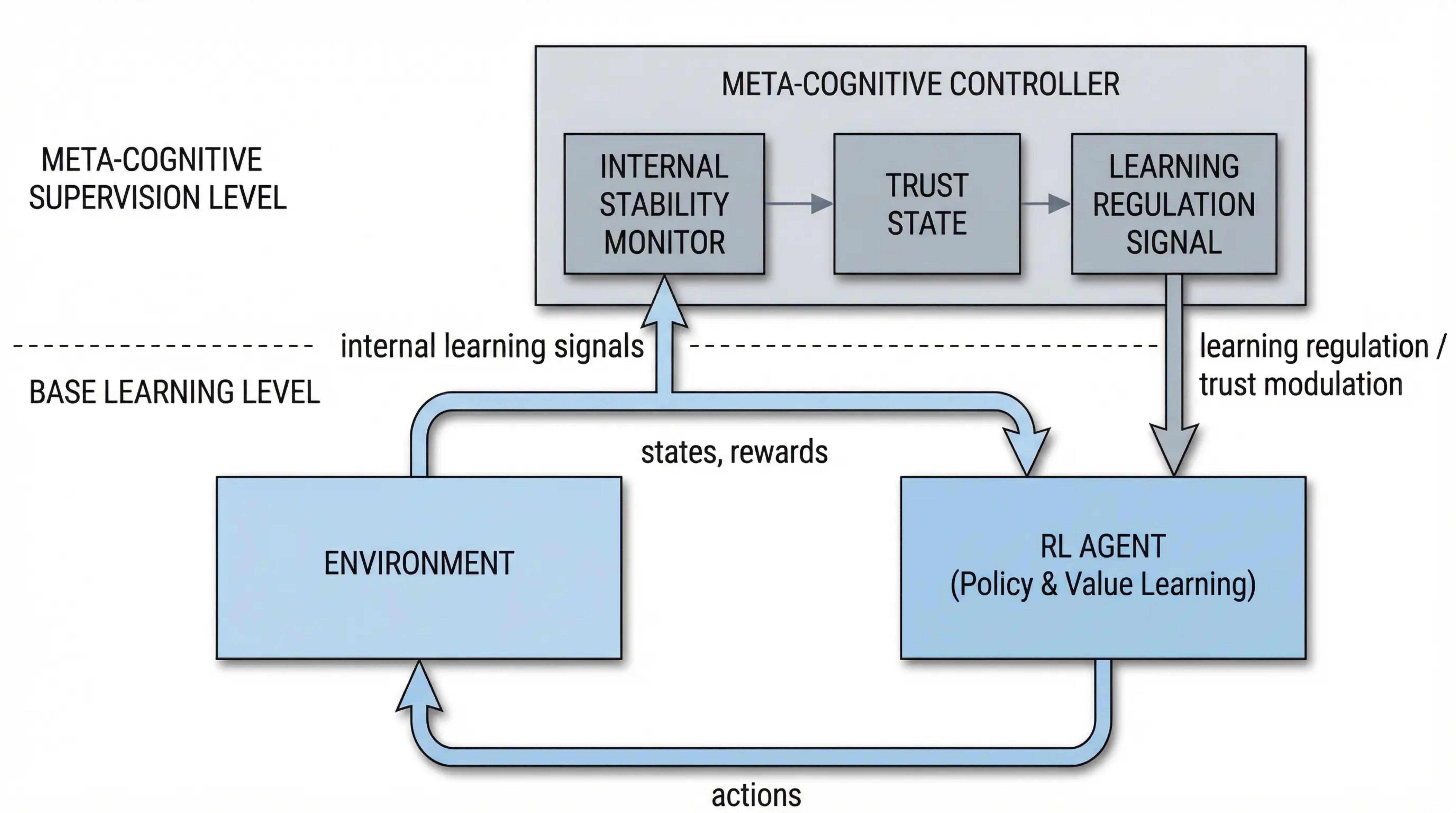}
    \caption{System-level architecture of the proposed meta-cognitive reinforcement learning framework.
    The agent is augmented with a meta-cognitive controller that monitors internal learning stability,
    maintains a trust state, and regulates learning dynamics through top-down supervision.}
    \label{fig:system_overview}
\end{figure}

Our approach augments a standard reinforcement learning agent with a supervisory meta-cognitive controller that operates on a slower time scale. As illustrated in Figure~\ref{fig:system_overview}, the controller monitors internal learning signals, maintains an explicit trust state, and modulates learning dynamics through top-down regulation. This modular design allows the framework to be integrated as a stabilizing component in existing RL-based control and decision systems, similar to how supervisory control layers are employed in industrial automation and robotics \cite{brunke2022safe, kiran2021deep}.

The two-timescale architecture—with fast policy updates and slower meta-cognitive regulation—draws inspiration from hierarchical control systems \cite{khalil2002nonlinear} and reflects practical considerations for real-time learning systems. By separating learning execution from learning regulation, the framework maintains compatibility with various base RL algorithms while providing a unified interface for stability management.

\section{Meta-Cognitive Reinforcement Learning Framework}\label{sec:framework}

\subsection{Overview of the Meta-Cognitive Control Framework}
We present a meta‑cognitive reinforcement learning framework that enables an agent to regulate its learning dynamics based on internal stability signals. \textbf{Our core proposition is that robust learning under uncertainty requires not just smarter optimization (``how to learn'') but explicit regulation of learning permission (``whether to learn'').} 

We consider a standard reinforcement learning setting modeled as a Markov Decision Process (MDP) defined by states $s \in \mathcal{S}$, actions $a \in \mathcal{A}$, transition dynamics $P$, reward function $r$, and discount factor $\gamma$. The agent learns a stochastic policy $\pi_\theta(a \mid s)$ parameterized by $\theta$ to maximize the expected return. Our implementation builds upon Proximal Policy Optimization (PPO) \cite{schulman2017ppo}, but the framework is compatible with other actor-critic methods, like TRPO \cite{schulman2015trpo}, SAC \cite{haarnoja2018sac}, or TD3 \cite{fujimoto2018td3}.

The framework augments a standard RL agent with a supervisory meta‑cognitive controller operating on a slower time scale, maintaining an explicit trust state that governs when learning should be suppressed or restored based on internal stability signals.

\paragraph{Conceptual Scope of Meta-Cognitive Control}
Unlike adaptive optimization methods that react to instantaneous gradient statistics or short-term reward signals,
the proposed meta-cognitive controller operates at a higher temporal abstraction level.
Rather than continuously tuning learning rates to accelerate convergence,
it regulates whether learning should proceed at all, based on accumulated evidence of internal stability.
This distinction positions meta-cognitive control as a process-level regulator of learning reliability,
conceptually complementary to robust reinforcement learning and adaptive optimization,
but not reducible to either.

\subsection{Value Prediction Error Stability (VPES)}
To assess the reliability of the agent's current learning dynamics, we introduce \emph{Value Prediction Error Stability} (VPES), a scalar signal that measures the temporal consistency of value estimation errors. Let $\delta_t$ denote the temporal-difference (TD) error at time step $t$. VPES is defined as the variance of TD errors over a sliding window of recent updates:
\begin{equation}
\mathrm{VPES}_t = \mathrm{Var}\left( \{ \delta_{t-k}, \dots, \delta_t \} \right).
\end{equation}

\textbf{Note: VPES is not intended as a formal Lyapunov function, but rather as a practical internal indicator of stability trends in approximate reinforcement learning systems.} In practice, VPES is used as a relative trend indicator rather than an absolute stability measure, and its role is to capture changes in internal consistency over time.
High VPES indicates inconsistent value predictions and internal instability, while low VPES suggests stable learning dynamics. Unlike reward-based corruption detectors, VPES depends solely on the agent's internal predictions, making it applicable even when external feedback is unreliable \cite{fedus2020revisiting, kumar2020discor}.

\subsection{Meta-Trust Dynamics and Fail-Safe Regulation}
We introduce a meta-trust variable $\tau_t \in [0,1]$ that represents the agent's confidence in its current learning process. To capture stability trends, we maintain an exponential moving average of VPES:
\begin{equation}
\bar{v}_t = (1 - \beta_v)\bar{v}_{t-1} + \beta_v \mathrm{VPES}_t,
\end{equation}
and define the stability trend as:
\begin{equation}
\Delta v_t = \bar{v}_t - \mathrm{VPES}_t.
\end{equation}

A positive $\Delta v_t$ indicates improving stability, while a negative value indicates worsening instability. Meta-trust is then updated asymmetrically:
\begin{equation}
\tau_t =
\begin{cases}
\min(1, \tau_{t-1} + \eta_{\text{up}}), & \text{if } \Delta v_t > 0, \\
\max(0, \tau_{t-1} - \eta_{\text{down}}), & \text{otherwise},
\end{cases}
\end{equation}
where $\eta_{\text{down}} > \eta_{\text{up}}$ ensures that trust decreases rapidly under instability but recovers gradually as stability returns. This asymmetric update reflects a cognitive principle: confidence is easier to lose than to regain \cite{nelson1990metamemory}.

Meta-trust modulates learning dynamics through a control signal that scales the effective learning rate:
\begin{equation}
\alpha_t = \alpha_0 \cdot c_t,
\end{equation}
where $\alpha_0$ is the base learning rate, and $c_t$ is derived from $\tau_t$.

To prevent catastrophic updates during periods of low confidence, we introduce a fail-safe constraint:
\begin{equation}
c_t \leq 1 \quad \text{if } \tau_t < \tau_{\text{min}}.
\end{equation}
This constraint ensures that when trust is low, updates may be reduced but never amplified, preventing runaway instability \cite{schulman2015trpo, achiam2017constrained}.

\subsection{Recovery Mechanism and Trust Restoration}
The recovery mechanism is embedded within the asymmetric update rule. When stability improves ($\Delta v_t > 0$), meta-trust slowly increases at rate $\eta_{\text{up}}$. This gradual restoration ensures that the agent cautiously resumes learning capacity only after sustained evidence of stability, preventing premature reactivation that could re-trigger instability.

The recovery process embodies a risk-sensitive control principle: it requires more evidence to regain trust than to lose it. This design is inspired by cognitive models of confidence restoration \cite{nelson1990metamemory} and aligns with safe learning practices in robotics \cite{brunke2022safe} and autonomous systems \cite{kiran2021deep}. As trust recovers, the fail-safe constraint is gradually relaxed, allowing the agent to resume learning at full capacity when stability is consistently maintained.

\subsection{Algorithmic Implementation}
\begin{figure}[htbp]
    \centering
    \includegraphics[width=0.85\linewidth]{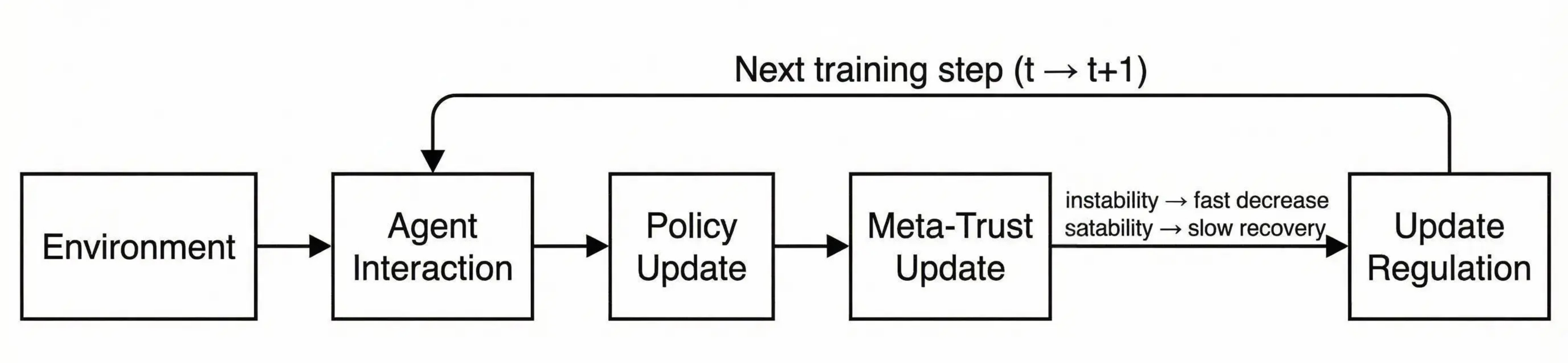}
    \caption{Overall training loop with asymmetric meta-cognitive control.
    The figure illustrates the execution flow of a single training iteration,
    where policy updates are regulated by meta-trust and repeated across
    iterations ($t \rightarrow t{+}1$).}
    \label{fig:training_loop}
\end{figure}

The complete algorithm integrates experience-level robustness and meta-cognitive control in a unified training loop. Figure~\ref{fig:training_loop} illustrates the overall workflow, and Algorithm~\ref{alg:meta_cognitive_rl} provides the pseudocode.

\begin{algorithm}[t]
\footnotesize
\caption{Meta-Cognitive Reinforcement Learning with Trust Recovery}
\label{alg:meta_cognitive_rl}
\begin{algorithmic}[1]

\Require Base learning rate $\alpha_0$,
         trust update rates $\eta_{\text{up}}, \eta_{\text{down}}$,
         VPES smoothing factor $\beta_v$,
         trust threshold $\tau_{\min}$

\State Initialize policy parameters $\theta$
\State Initialize meta-trust $\tau_0 \in (0,1)$
\State Initialize VPES baseline $\bar{v}_0 \gets 0$

\For{each training iteration $t = 1,2,\dots$}

    \State Collect trajectories using policy $\pi_\theta$
    \State Compute TD errors $\{\delta_i\}$ from collected data

    \State Compute VPES:
    \[
        \mathrm{VPES}_t \gets \mathrm{Var}(\{\delta_i\})
    \]

    \State Update VPES baseline:
    \[
        \bar{v}_t \gets (1-\beta_v)\bar{v}_{t-1} + \beta_v \mathrm{VPES}_t
    \]

    \State Compute stability trend $\Delta v_t \gets \bar{v}_t - \mathrm{VPES}_t$

    \If{$\Delta v_t > 0$} \Comment{stability improving}
        \State $\tau_t \gets \min\{1,\; \tau_{t-1} + \eta_{\text{up}}\}$
    \Else \Comment{stability deteriorating}
        \State $\tau_t \gets \max\{0,\; \tau_{t-1} - \eta_{\text{down}}\}$
    \EndIf

    \State Compute learning-rate scale $c_t \gets f(\tau_t)$

    \If{$\tau_t < \tau_{\min}$} \Comment{fail-safe regulation}
        \State $c_t \gets \min\{c_t,\; 1.0\}$
    \EndIf

    \State Set effective learning rate $\alpha_t \gets \alpha_0 \cdot c_t$

    \State Update policy parameters $\theta$ using PPO with learning rate $\alpha_t$

\EndFor

\end{algorithmic}
\end{algorithm}

The algorithm operates at two time scales: a primary learning loop (policy updates) and a slower meta-cognitive control loop (trust updates). This separation maintains compatibility with various base RL algorithms while providing a unified interface for stability management. The framework can be integrated as a stabilizing component in existing RL systems without requiring extensive architectural changes.

\section{Experiments}

This section evaluates the proposed VPES control mechanism under
controlled training collapse scenarios. Rather than focusing solely on
asymptotic performance, the experiments investigate how learning
dynamics behave after collapse events and whether adaptive control
improves recovery.

We address the following questions:

\begin{itemize}
\item Can collapse be reproducibly induced through a controlled perturbation protocol?
\item Does VPES improve recovery behaviour after collapse?
\item Do collapse events exhibit distinct regimes?
\item How does VPES respond internally to collapse signals?
\item Is adaptive learning-rate control necessary compared with fixed decay strategies?
\item Is VPES robust to different perturbation strengths?
\end{itemize}

\subsection{Collapse–Recovery Experimental Protocol}

Experiments follow a three-phase protocol.

\paragraph{Phase 1: checkpoint selection}

A PPO agent is first trained normally. During training we record
multiple checkpoints with high evaluation return. These checkpoints
represent stable policy states along the training trajectory.

From the full training run we select 23 checkpoints as starting points
for collapse experiments.

\paragraph{Phase 2: internal perturbation}

Collapse is induced through internal perturbations applied directly to
the learning signal. Specifically, the sign of the advantage estimate
is flipped with probability $p$:

\begin{equation}
A_t \leftarrow
\begin{cases}
-A_t & \text{with probability } p \\
A_t & \text{otherwise}
\end{cases}
\end{equation}

This perturbation alters the policy gradient direction while leaving
the environment unchanged. As a result, collapse events can be
introduced in a controlled manner.

\paragraph{Phase 3: recovery}

After perturbation ends, training continues without further
intervention. Recovery behaviour is then observed.

All methods start from the same checkpoint so that differences in
recovery dynamics are attributable only to the learning algorithm.

\subsection{Collapse Generation}

We first verify that the perturbation protocol reliably produces
collapse events.

For each checkpoint we record the maximum return before perturbation
($PrePerturbPeak$) and the minimum return observed during perturbation
($ShockMin$). Collapse depth is defined as

\begin{equation}
CollapseDepth = PrePerturbPeak - ShockMin .
\end{equation}

To normalize across checkpoints with different performance scales we
define

\begin{equation}
CollapseRatio = \frac{CollapseDepth}{PrePerturbPeak}.
\end{equation}

Across the 23 checkpoints the perturbation produces collapses of
varying severity, ranging from mild performance degradation to
catastrophic failure. This variation allows recovery behaviour to be
studied under different collapse conditions.

\subsection{Collapse–Recovery Dynamics}

\begin{figure}[t]
\centering
\includegraphics[width=\linewidth]{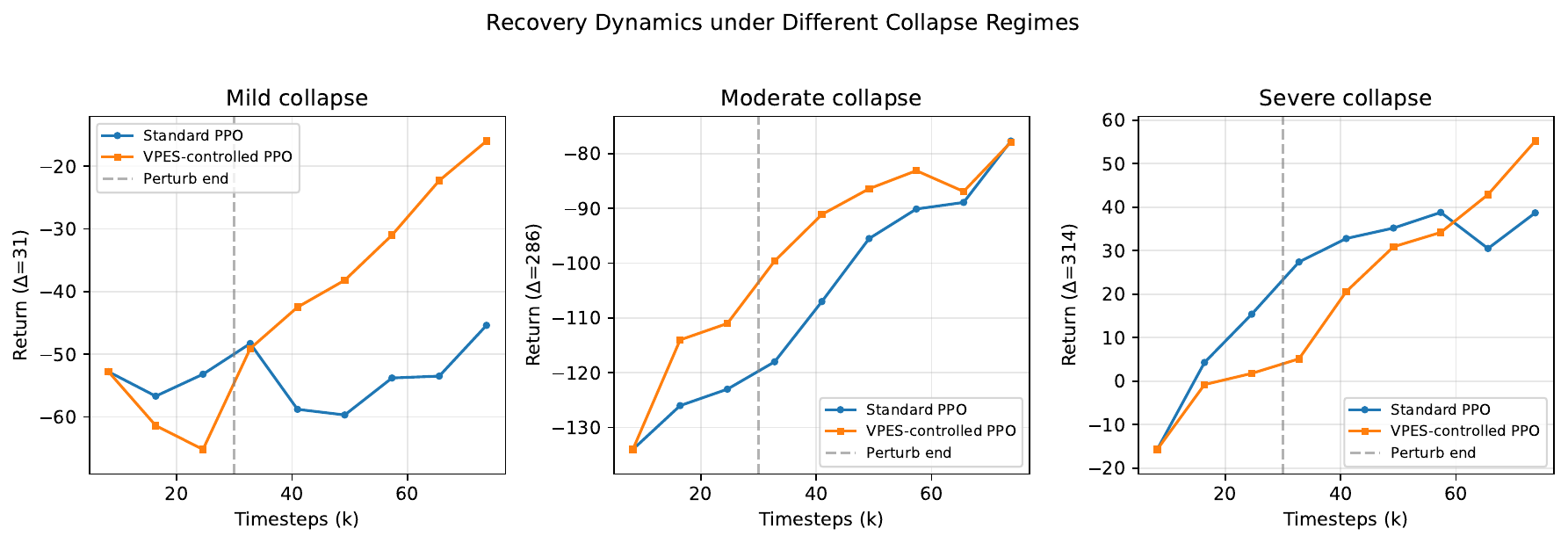}
\caption{
Collapse–recovery trajectories under three representative regimes.
Each subplot shows the evaluation return during and after
perturbation for standard PPO and PPO with VPES control.
The dashed line marks the end of the perturbation phase.
}
\label{fig:collapse_recovery}
\end{figure}

Representative training trajectories are shown in
Figure~\ref{fig:collapse_recovery}.

In mild collapse cases both methods recover quickly with little
difference.

In moderate collapse regimes VPES typically recovers earlier and
maintains higher return during the recovery phase.

In severe collapse regimes recovery becomes more difficult for both
methods, although VPES sometimes reduces the depth of collapse.

These observations suggest that the main effect of VPES appears in the
recovery phase rather than in final performance.

\subsection{Mechanism Analysis}

To examine how VPES reacts to collapse signals we analyze its internal
control behaviour.

VPES monitors the relative change in rollout return between
consecutive updates:

\begin{equation}
g_t = \frac{R_t - R_{t-1}}{|R_{t-1}| + \epsilon}.
\end{equation}

When the return drop exceeds a threshold the learning rate is reduced:

\begin{equation}
lr\_scale \leftarrow lr\_scale (1 - \eta_{down})
\end{equation}

where $\eta_{down}=0.15$.

\begin{figure}[t]
\centering

\begin{subfigure}{0.9\linewidth}
\centering
\includegraphics[width=\linewidth]{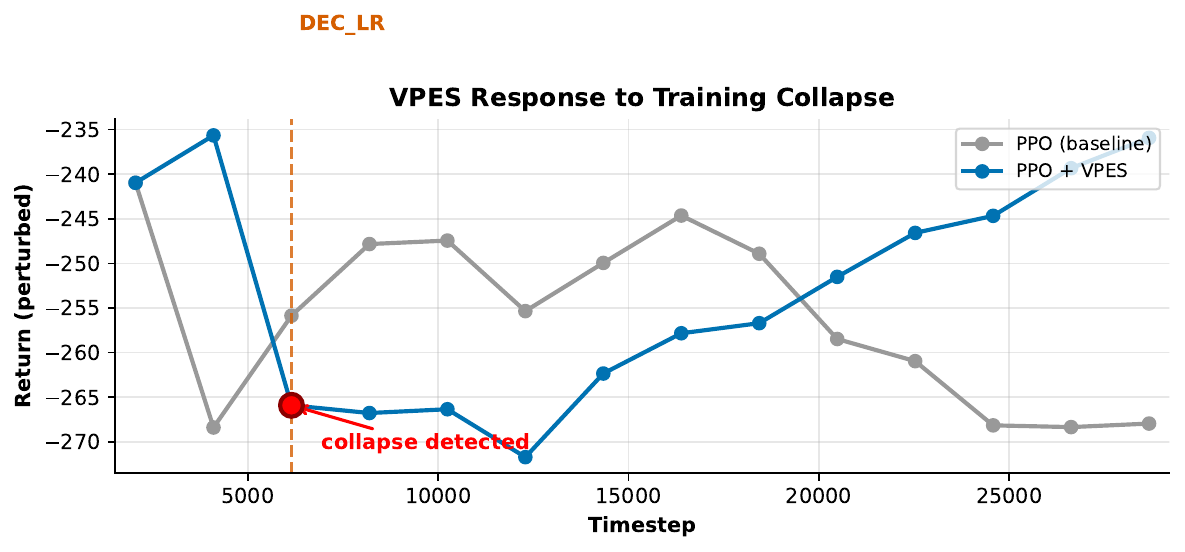}
\caption{Return trajectories for PPO and PPO+VPES.}
\end{subfigure}

\vspace{0.4cm}

\begin{subfigure}{0.9\linewidth}
\centering
\includegraphics[width=\linewidth]{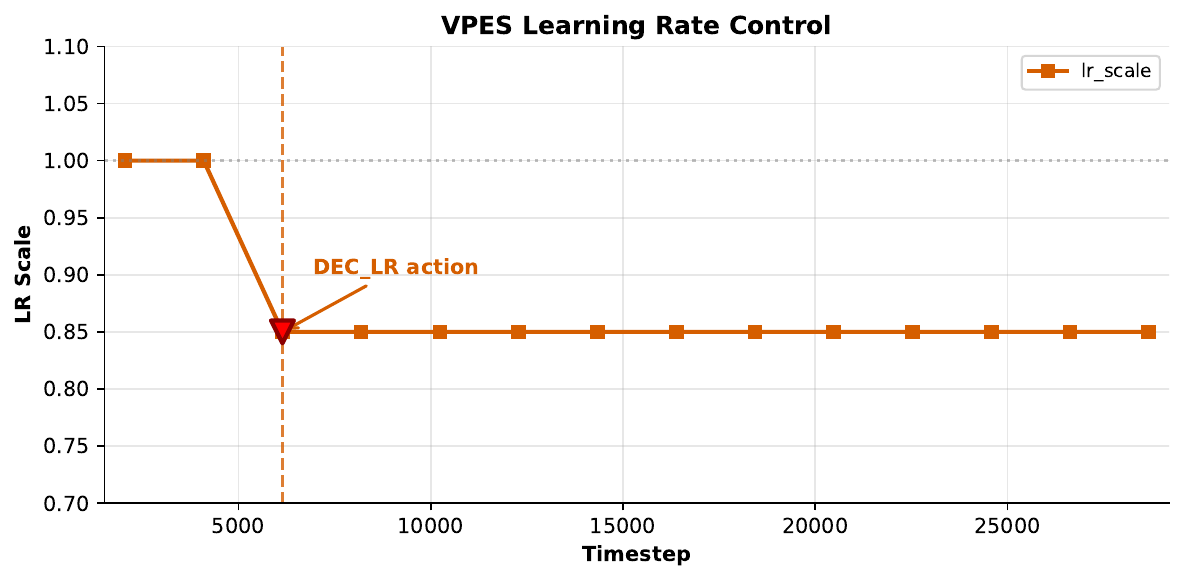}
\caption{Learning-rate control signal produced by VPES.}
\end{subfigure}

\caption{
Response of VPES during a collapse event.
(a) Return trajectories under identical perturbation conditions.
VPES recovers earlier than PPO.
(b) Corresponding learning-rate control signal.
}
\label{fig:vpes_mechanism}

\end{figure}

Figure~\ref{fig:vpes_mechanism} illustrates a representative collapse
event from checkpoint\_23.

Figure~\ref{fig:vpes_mechanism}(a) compares return trajectories for PPO
and PPO+VPES under the same perturbation. While both methods experience
collapse, the VPES-controlled run begins recovering earlier.

Figure~\ref{fig:vpes_mechanism}(b) shows the corresponding learning-rate
control signal. The learning rate is temporarily reduced after collapse
detection, stabilizing the subsequent updates.

\subsection{Overall Comparison}

Table~\ref{tab:overall_results} summarizes results across all
23 checkpoints.

\begin{table}[t]
\centering
\caption{Overall comparison across 23 checkpoints.}
\label{tab:overall_results}
\begin{tabular}{lcc}
\toprule
Metric & VPES Wins & PPO Wins \\
\midrule
Final Return & 13 & 10 \\
Recovery Efficiency & 12 & 11 \\
\bottomrule
\end{tabular}
\end{table}

VPES achieves slightly higher win rates on both metrics.

\subsection{Collapse Regimes}

Collapse events are categorized into three regimes according to
collapse severity:

\begin{center}
\begin{tabular}{cc}
\toprule
Regime & CollapseRatio \\
\midrule
Mild & $<0.3$ \\
Moderate & $0.3-0.5$ \\
Severe & $\ge0.5$ \\
\bottomrule
\end{tabular}
\end{center}

Results are summarized in Table~\ref{tab:regime_analysis}.

\begin{table}[t]
\centering
\caption{VPES win rates under different collapse regimes.}
\label{tab:regime_analysis}
\begin{tabular}{lccc}
\toprule
Regime & Seeds & Final Win Rate & RecEff Win Rate \\
\midrule
Mild & 11 & 55\% & 36\% \\
Moderate & 8 & 63\% & 75\% \\
Severe & 4 & 50\% & 50\% \\
\bottomrule
\end{tabular}
\end{table}

The advantage of VPES is most visible in the moderate collapse regime,
where instability is present but recovery remains possible.

\subsection{Collapse Severity and Recovery}

We further examine the relationship between collapse severity and
recovery improvement.

\begin{table}[t]
\centering
\caption{Correlation between collapse severity and recovery improvement.}
\label{tab:correlation}
\begin{tabular}{lc}
\toprule
Analysis & Correlation \\
\midrule
CollapseRatio $\rightarrow$ Final Improvement & 0.19 \\
CollapseRatio $\rightarrow$ RecEff Improvement & 0.29 \\
\bottomrule
\end{tabular}
\end{table}

Stronger collapses tend to correspond to larger improvements in
recovery efficiency.

\subsection{Failure Cases}

In some cases collapse becomes irreversible. For example,
checkpoint\_23 exhibits a very deep collapse after perturbation,
after which recovery does not occur within the available training
horizon.

These cases indicate the presence of catastrophic collapse regimes
where the policy diverges too far from the original attractor.

\subsection{Ablation: Learning-Rate Control}

We compare VPES with a fixed learning-rate decay strategy.

Methods include:

\begin{itemize}
\item PPO
\item PPO + Fixed LR Decay
\item PPO + VPES
\end{itemize}

Experiments are conducted on three representative checkpoints.

Fixed decay can stabilize training in some cases but requires manual
tuning and behaves inconsistently across collapse regimes. VPES
adjusts the learning rate dynamically and provides more consistent
behaviour across conditions.

\subsection{Robustness to Perturbation Strength}

Finally we vary perturbation strength by changing the advantage flip
probability $p \in \{0.3,0.5,0.7\}$.

Across representative checkpoints VPES maintains similar recovery
behaviour under moderate and strong perturbations, indicating that the
control mechanism does not rely on a specific perturbation strength.

\subsection{Summary}

The experiments show that

\begin{itemize}
\item collapse events can be reliably induced through internal perturbations,
\item VPES improves recovery dynamics particularly in moderate collapse regimes,
\item the learning-rate controller reacts directly to collapse signals during training.
\end{itemize}

\section{Discussion}\label{sec:discussion}

\subsection{From Robustness Heuristics to Meta-Cognitive Control}
The experimental results highlight a fundamental distinction between robustness-oriented learning and meta-cognitive control.
Conventional robust reinforcement learning methods primarily focus on filtering unreliable experiences or constraining updates based on external signals.
While effective in mitigating noise, these approaches implicitly assume that learning should proceed whenever sufficient data is available.

In contrast, our framework treats learning itself as an object of regulation.
By introducing meta-trust as an explicit cognitive state, the agent reasons about the reliability of its own learning dynamics and adjusts its behavior accordingly.
This shift—from suppressing unreliable data to regulating when and how learning should occur—marks a conceptual step toward meta-cognitive reinforcement learning systems.

Importantly, meta-cognitive control does not replace existing robustness heuristics, but complements them by addressing a distinct class of failure modes.
While robustness techniques focus on data quality and local update stability, meta-cognitive regulation operates at the level of learning-process reliability, enabling the agent to suspend, attenuate, or resume learning based on internal assessments of stability.
This complementary relationship helps explain why meta-cognitive control remains effective even when combined with strong optimization-focused baselines.

\subsection{Practical Implications for Long-Horizon and Safety-Critical Systems}
The meta-cognitive framework presented here is not merely an algorithmic improvement but a step toward more resilient and self-regulating autonomous systems.
In practical deployments, engineers often face a dilemma: overly aggressive online learning can lead to catastrophic failure, while disabling learning entirely forfeits adaptation.
Our method offers a principled middle ground by allowing learning intensity to be modulated rather than switched on or off.

For instance, in industrial predictive maintenance, a reinforcement learning agent optimizing maintenance schedules may receive corrupted reward signals due to faulty sensors.
A conventional robust RL method might persistently down-weight these signals, gradually degrading performance.
In contrast, a meta-cognitively equipped agent would detect the inconsistency via VPES, temporarily reduce its learning rate (entering a ``cautious'' mode), and only resume full learning capacity when the sensor readings return to a stable pattern.

Similarly, in autonomous vehicle platooning, where vehicles learn cooperative driving policies via reinforcement learning, communication delays or adversarial interference can create non-stationary reward channels.
A meta-cognitive controller could allow individual vehicles to autonomously scale back policy updates during unstable communication phases, preventing the propagation of erroneous learning across the platoon, and then recover coordination once channel quality improves.

\subsection{Interpretability and Cognitive Plausibility}
A notable advantage of the proposed framework is its interpretability.
The meta-trust variable has a clear semantic meaning: it reflects the agent's confidence in its current learning process.
The asymmetric update rule—rapid loss of trust under instability and slow recovery under improving stability—aligns with well-established cognitive principles and provides an intuitive explanation for the observed learning dynamics.

Moreover, all signals used by the meta-cognitive controller are internally generated.
VPES depends solely on the agent's value predictions and does not rely on external annotations or assumptions about reward corruption.
This internal consistency not only enhances conceptual clarity, but also facilitates debugging, monitoring, and system-level reasoning in practical deployments where external reliability indicators may be unavailable or unreliable.

\subsection{Trade-Offs Between Performance and Tail Risk}
Our results reveal an explicit trade-off between average performance and tail risk.
While the recovery-enabled meta-cognitive controller improves mean returns and significantly reduces late-stage failures, its CVaR@20\% may remain lower than that of some aggressive optimization baselines.
Importantly, this behavior is not indicative of algorithmic deficiency, but rather reflects a deliberate design choice: when internal stability remains ambiguous, the agent prioritizes caution over short-term performance gains.

From a system engineering perspective, increased variance under meta-cognitive control represents a conscious trade-off rather than a weakness.
By allowing heterogeneous recovery dynamics across runs, the agent avoids enforcing overly rigid learning schedules.
This flexibility increases dispersion among successful trajectories while substantially reducing the probability of catastrophic failure.
Such a trade-off is often preferable in long-horizon and safety-critical applications, where avoiding collapse outweighs minimizing variance across outcomes.

\subsection{Limitations and Future Directions}
Despite its advantages, the proposed framework has several limitations.
First, the current implementation relies on manually specified update rates for trust recovery and degradation.
While these parameters were kept fixed across experiments and exhibited reasonable cross-environment robustness, adaptive or learned recovery rates may further improve performance and reduce tail risk.

Second, our experiments focus on continuous-control benchmarks with synthetic reward corruption.
Although this setting effectively exposes late-stage instability, broader validation across additional environments, corruption mechanisms, and real-world noise sources is necessary to fully characterize the generality of the approach.

Finally, the meta-cognitive control mechanism currently operates at the level of learning-rate modulation.
Extending the framework to regulate other aspects of learning—such as exploration strategies, experience replay selection, or architectural adaptation—remains an open and promising direction for future research.

\section{Conclusion}

This paper presents a meta-cognitive reinforcement learning framework that enables agents to autonomously regulate their own learning dynamics under unreliable and non-stationary feedback.
By introducing \emph{meta-trust} as an explicit internal cognitive state, driven by Value Prediction Error Stability (VPES), the proposed method allows the learning process itself to become an object of regulation.
Rather than continuously optimizing in the presence of uncertain signals, the agent can detect internal instability, suppress risky updates, and gradually restore learning capacity once stable dynamics re-emerge.
This represents a conceptual shift from conventional robustness heuristics toward learning-aware, self-regulating control.
By explicitly separating learning execution from learning permission, the proposed framework provides a system-level abstraction for managing learning reliability that is orthogonal to algorithmic improvements in policy optimization.

Extensive experimental evaluations demonstrate that the proposed framework substantially reduces late-stage training failures under reward corruption and non-stationary disturbances.
While improvements in average return are observed in several settings, the primary empirical benefit lies in mitigating tail risk and preventing catastrophic learning collapse.
Across stationary, non-stationary, and cross-environment scenarios, asymmetric trust recovery consistently stabilizes long-horizon training dynamics, confirming that the method improves reliability rather than merely optimizing short-term performance.
Ablation studies further show that asymmetric recovery is a necessary control mechanism to avoid low-trust stagnation and ensure sustained adaptation.

From a systems perspective, the proposed meta-cognitive control framework offers a practical solution for reinforcement learning in long-horizon and safety-critical decision-making scenarios, where uncontrolled online learning may lead to irreversible failure.
By regulating \emph{when} and \emph{how} learning should proceed based on internal stability signals, the framework provides a principled balance between adaptation and risk control.
These findings suggest that meta-cognitive regulation constitutes a promising direction for building resilient, self-monitoring learning systems in real-world applications characterized by uncertainty and delayed feedback.

% ==================== References ====================
\bibliographystyle{elsarticle-num}
\bibliography{refs}

@article{abdar2021review,
  title = {A Review of Uncertainty Quantification in Deep Learning: {{Techniques}}, Applications and Challenges},
  author = {Abdar, Moloud and Pourpanah, Farhad and Hussain, Sadiq and Rezazadegan, Dana and Liu, Li and Ghavamzadeh, Mohammad and Fieguth, Paul and Cao, Xiaochun and Khosravi, Abbas and Acharya, U. Rajendra and Makarenkov, Vladimir and Nahavandi, Saeid},
  year = 2021,
  journal = {Information Fusion},
  volume = {76},
  pages = {243--297},
  issn = {1566-2535},
  doi = {10.1016/j.inffus.2021.05.008}
}

@inproceedings{achiam2017constrained,
  title = {Constrained Policy Optimization},
  booktitle = {Proceedings of the International Conference on Machine Learning},
  author = {Achiam, Joshua and Held, David and Tamar, Aviv and Abbeel, Pieter},
  year = 2017,
  pages = {22--31}
}

@inproceedings{al2018continuous,
  title = {Continuous Adaptation via Meta-Learning in Non-Stationary and Competitive Environments},
  booktitle = {{{ICLR}}},
  author = {{Al-Shedivat}, Maruan and Bansal, Trapit and Burda, Yuri and Sutskever, Ilya and Mordatch, Igor and Abbeel, Pieter},
  year = 2018,
  pages = {1--}
}

@article{brunke2022safe,
  title = {Safe Learning in Robotics: {{From}} Learning-Based Control to Safe Reinforcement Learning},
  author = {Brunke, Lukas and Greeff, Melissa and Hall, Adam W and Yuan, Zhaocong and Zhou, Siqi and Panerati, Jacopo and Schoellig, Angela P},
  year = 2022,
  journal = {Annual Review of Control, Robotics, and Autonomous Systems},
  volume = {5},
  pages = {269--297},
  publisher = {Annual Reviews}
}

@inproceedings{dalal2018safe,
  title = {Safe Exploration in Continuous Action Spaces},
  booktitle = {{{AAAI}}},
  author = {Dalal, Gal and Dvijotham, Krishnamurthy and Vecerik, Matej and Hester, Todd and Paduraru, Cosmin and Tassa, Yuval},
  pages = {1--},
  year = 2018
}

@misc{engstrom2020impl,
  title = {Implementation Matters in Deep Policy Gradients: A Case Study on {{PPO}} and {{TRPO}}},
  author = {Engstrom, Logan and Ilyas, Andrew and Santurkar, Shibani and Tsipras, Dimitris and Janoos, Firdaus and Rudolph, Larry and Madry, Aleksander},
  year = 2020,
  eprint = {2005.12729},
  primaryclass = {cs.LG},
  archiveprefix = {arXiv}
}

@inproceedings{everitt2017reinforcement,
  title = {Reinforcement Learning with a Corrupted Reward Channel},
  booktitle = {Proceedings of the 26th International Joint Conference on Artificial Intelligence},
  author = {Everitt, Tom and Krakovna, Victoria and Orseau, Laurent and Legg, Shane},
  year = 2017,
  series = {{{IJCAI}}'17},
  pages = {4705--4713},
  publisher = {AAAI Press},
  address = {Melbourne, Australia}
}

@inproceedings{fedus2020revisiting,
  title = {Revisiting Fundamentals of Experience Replay},
  booktitle = {Proceedings of the International Conference on Machine Learning},
  author = {Fedus, William and Ramachandran, Prajit and Agarwal, Rishabh and Bengio, Yoshua},
  year = 2020,
  pages = {3061--3071}
}

@article{flavell1979metacognition,
  title = {Metacognition and Cognitive Monitoring: {{A}} New Area of Cognitive-Developmental Inquiry},
  author = {Flavell, John H},
  year = 1979,
  journal = {American Psychologist},
  volume = {34},
  number = {10},
  pages = {906}
}

@inproceedings{fujimoto2018td3,
  title = {Addressing Function Approximation Error in Actor-Critic Methods},
  booktitle = {Proceedings of the International Conference on Machine Learning},
  author = {Fujimoto, Scott and Hoof, Herke and Meger, David},
  year = 2018,
  pages = {1587--1596}
}

@inproceedings{fujimoto2019off,
  title = {Off-Policy Deep Reinforcement Learning without Exploration},
  booktitle = {Proceedings of the International Conference on Machine Learning},
  author = {Fujimoto, Scott and Meger, David and Precup, Doina},
  year = 2019,
  pages = {2052--2062}
}

@inproceedings{gal2016dropout,
  title = {Dropout as a Bayesian Approximation: {{Representing}} Model Uncertainty in Deep Learning},
  booktitle = {Proceedings of the International Conference on Machine Learning},
  author = {Gal, Yarin and Ghahramani, Zoubin},
  editor = {Balcan, Maria Florina and Weinberger, Kilian Q.},
  year = 2016,
  month = jun,
  series = {Proceedings of Machine Learning Research},
  volume = {48},
  pages = {1050--1059},
  publisher = {PMLR},
  address = {New York, New York, USA}
}

@article{garcia2015comprehensive,
  title = {A Comprehensive Survey on Safe Reinforcement Learning},
  author = {Garcia, Javier and Fern{\'a}ndez, Fernando},
  year = 2015,
  journal = {Journal of Machine Learning Research},
  volume = {16},
  number = {1},
  pages = {1437--1480}
}

@article{ghavamzadeh2015bayesian,
  title = {Bayesian Reinforcement Learning: A Survey},
  author = {Ghavamzadeh, Mohammad and Mannor, Shie and Pineau, Joelle and Tamar, Aviv},
  year = 2015,
  month = nov,
  journal = {Found. Trends Mach. Learn.},
  volume = {8},
  number = {5--6},
  pages = {359--483},
  publisher = {Now Publishers Inc.},
  address = {Hanover, MA, USA},
  issn = {1935-8237},
  doi = {10.1561/2200000049},
  issue_date = {Nov 2015}
}

@article{griffiths2019doing,
  title = {Doing More with Less: Meta-Reasoning and Meta-Learning in Humans and Machines},
  author = {Griffiths, Thomas L and Callaway, Frederick and Chang, Michael B and Grant, Erin and Krueger, Paul M and Lieder, Falk},
  year = 2019,
  journal = {Current Opinion in Behavioral Sciences},
  volume = {29},
  pages = {24--30},
  issn = {2352-1546},
  doi = {10.1016/j.cobeha.2019.01.005}
}

@article{gu2024review,
  title = {A Review of Safe Reinforcement Learning: {{Methods}}, Theories, and Applications},
  author = {Gu, Shangding and Yang, Long and Du, Yali and Chen, Guang and Walter, Florian and Wang, Jun and Knoll, Alois},
  year = 2024,
  journal = {IEEE Transactions on Pattern Analysis and Machine Intelligence},
  volume = {46},
  number = {12},
  pages = {11216--11235},
  doi = {10.1109/TPAMI.2024.3457538}
}

@inproceedings{haarnoja2018sac,
  title = {Soft Actor-Critic: {{Off-policy}} Maximum Entropy Deep Reinforcement Learning with a Stochastic Actor},
  booktitle = {Proceedings of the International Conference on Machine Learning},
  author = {Haarnoja, Tuomas and Zhou, Aurick and Abbeel, Pieter and Levine, Sergey},
  year = 2018,
  pages = {1861--1870}
}

@inproceedings{henderson2018deep,
  title = {Deep Reinforcement Learning That Matters},
  booktitle = {{{AAAI}}},
  author = {Henderson, Peter and Islam, Riashat and Bachman, Philip and Pineau, Joelle and Precup, Doina and Meger, David},
  year = 2018,
  series = {{{AAAI}}'18/{{IAAI}}'18/{{EAAI}}'18},
  publisher = {AAAI Press},
  address = {New Orleans, Louisiana, USA},
  articleno = {392},
  pages = {1--},
  isbn = {978-1-57735-800-8}
}

@book{khalil2002nonlinear,
  title = {Nonlinear Systems},
  author = {Khalil, Hassan K.},
  year = 2002,
  edition = {3rd},
  publisher = {Prentice Hall}
}

@article{kiran2021deep,
  title = {Deep Reinforcement Learning for Autonomous Driving: A Survey},
  author = {Kiran, B Ravi and Sobh, Ibrahim and Talpaert, Victor and Mannion, Patrick and Sallab, Ahmad A. Al and Yogamani, Senthil and P{\'e}rez, Patrick},
  year = 2022,
  journal = {IEEE Transactions on Intelligent Transportation Systems},
  volume = {23},
  number = {6},
  pages = {4909--4926},
  doi = {10.1109/TITS.2021.3054625}
}

@inproceedings{kumar2020conservative,
  title = {Conservative Q-Learning for Offline Reinforcement Learning},
  booktitle = {{{NeurIPS}}},
  author = {Kumar, Aviral and Zhou, Aurick and Tucker, George and Levine, Sergey},
  year = 2020,
  volume = {33},
  pages = {1179--1191}
}

@inproceedings{kumar2020discor,
  title = {{{DisCor}}: {{Corrective}} Feedback in Reinforcement Learning via Distribution Correction},
  booktitle = {{{NeurIPS}}},
  author = {Kumar, Aviral and Gupta, Abhishek and Levine, Sergey},
  year = 2020,
  volume = {33},
  pages = {18560--18572}
}

@article{lake2017building,
  title = {Building Machines That Learn and Think like People},
  author = {Lake, Brenden M and Ullman, Tomer D and Tenenbaum, Joshua B and Gershman, Samuel J},
  year = 2017,
  journal = {Behavioral and Brain Sciences},
  volume = {40},
  pages = {e253},
  publisher = {Cambridge University Press}
}

@misc{mankowitz2020robust,
  title = {Robust Reinforcement Learning for Continuous Control with Model Misspecification},
  author = {Mankowitz, Daniel J. and Levine, Nir and Jeong, Rae and Shi, Yuanyuan and Kay, Jackie and Abdolmaleki, Abbas and Springenberg, Jost Tobias and Mann, Timothy and Hester, Todd and Riedmiller, Martin},
  year = {2019},
  eprint = {1906.07516},
  archiveprefix = {arXiv},
  primaryclass = {cs.LG}
}

@article{moos2022robust,
  title = {Robust Reinforcement Learning: A Review of Foundations and Recent Advances},
  author = {Moos, Janosch and Hansel, Kay and Abdulsamad, Hany and Stark, Svenja and Clever, Debora and Peters, Jan},
  year = 2022,
  journal = {Machine Learning and Knowledge Extraction},
  volume = {4},
  number = {1},
  pages = {276--315},
  issn = {2504-4990},
  doi = {10.3390/make4010013}
}

@incollection{nelson1990metamemory,
  author    = {Nelson, Thomas O.},
  title     = {Metamemory: A Theoretical Framework and New Findings},
  booktitle = {Psychology of Learning and Motivation},
  editor    = {Bower, Gordon H.},
  volume    = {26},
  pages     = {125--173},
  year      = {1990},
  publisher = {Academic Press},
  doi       = {10.1016/S0079-7421(08)60053-5}
}

@article{niv2009reinforcement,
  title = {Reinforcement Learning in the Brain},
  author = {Niv, Yael},
  year = 2009,
  journal = {Journal of Mathematical Psychology},
  volume = {53},
  number = {3},
  pages = {139--154},
  issn = {0022-2496},
  doi = {10.1016/j.jmp.2008.12.005}
}

@inproceedings{osband2016deep,
  title = {Deep Exploration via Bootstrapped {{DQN}}},
  booktitle = {Advances in Neural Information Processing Systems},
  author = {Osband, Ian and Blundell, Charles and Pritzel, Alexander and Van Roy, Benjamin},
  editor = {Lee, D. and Sugiyama, M. and Luxburg, U. and Guyon, I. and Garnett, R.},
  year = 2016,
  volume = {29},
  pages = {1--},
  publisher = {Curran Associates, Inc.}
}

@inproceedings{park2020meta,
  title = {Meta-Curvature},
  booktitle = {{{NeurIPS}}},
  author = {Park, Eunbyung and Oliva, Junier},
  year = 2019,
  pages = {1--},
  volume = {33}
}

@inproceedings{pathak2019self,
  title = {Self-Supervised Exploration via Disagreement},
  booktitle = {Proceedings of the International Conference on Machine Learning},
  author = {Pathak, Deepak and Gandhi, Dhiraj and Gupta, Abhinav},
  year = 2019,
  pages = {5062--5071},
  publisher = {PMLR}
}

@inproceedings{pinto2017robust,
  title = {Robust Adversarial Reinforcement Learning},
  booktitle = {Proceedings of the International Conference on Machine Learning},
  author = {Pinto, Lerrel and Davidson, James and Sukthankar, Rahul and Gupta, Abhinav},
  year = 2017,
  pages = {2817--2826},
  publisher = {PMLR}
}

@inproceedings{rabinowitz2018machine,
  title = {Machine Theory of Mind},
  booktitle = {Proceedings of the 35th International Conference on Machine Learning},
  author = {Rabinowitz, Neil and Perbet, Frank and Song, Francis and Zhang, Chiyuan and Eslami, S. M. Ali and Botvinick, Matthew},
  editor = {Dy, Jennifer and Krause, Andreas},
  year = 2018,
  month = jul,
  series = {Proceedings of Machine Learning Research},
  volume = {80},
  pages = {4218--4227},
  publisher = {PMLR}
}

@inproceedings{schulman2015trpo,
  title = {Trust Region Policy Optimization},
  booktitle = {Proceedings of the International Conference on Machine Learning},
  author = {Schulman, John and Levine, Sergey and Abbeel, Pieter and Jordan, Michael and Moritz, Philipp},
  editor = {Bach, Francis and Blei, David},
  year = 2015,
  series = {Proceedings of Machine Learning Research},
  volume = {37},
  pages = {1889--1897},
  publisher = {PMLR},
  address = {Lille, France}
}

@misc{schulman2017ppo,
  title = {Proximal Policy Optimization Algorithms},
  author = {Schulman, John and Wolski, Filip and Dhariwal, Prafulla and Radford, Alec and Klimov, Oleg},
  year = {2017},
  eprint = {1707.06347},
  archiveprefix = {arXiv},
  primaryclass = {cs.LG}
}

@inproceedings{xu2018meta,
  title = {Meta-Gradient Reinforcement Learning},
  booktitle = {Advances in Neural Information Processing Systems},
  author = {Xu, Zhongwen and {van Hasselt}, Hado P and Silver, David},
  editor = {Bengio, S. and Wallach, H. and Larochelle, H. and Grauman, K. and {Cesa-Bianchi}, N. and Garnett, R.},
  year = 2018,
  volume = {31},
  pages = {1--},
  publisher = {Curran Associates, Inc.}
}

@inproceedings{zahavy2020self,
  title = {Self-Tuning Deep Reinforcement Learning},
  booktitle = {{{NeurIPS}}},
  author = {Zahavy, Tom and Xu, Zhongwen and Veeriah, Vivek and Hessel, Matteo and Oh, Junhyuk and {van Hasselt}, Hado and Silver, David and Singh, Satinder},
  year = 2020,
  volume = {34},
  pages = {11871--11882}
}

@inproceedings{zhang2020robust,
  title = {Robust Deep Reinforcement Learning against Adversarial Perturbations on State Observations},
  booktitle = {{{NeurIPS}}},
  author = {Zhang, Huan and Chen, Hongge and Xiao, Chaowei and Li, Bo and Liu, Mingyan and Boning, Duane and Hsieh, Cho-Jui},
  year = 2020,
  series = {Nips '20},
  publisher = {Curran Associates Inc.},
  address = {Red Hook, NY, USA},
  articleno = {1765},
  pages = {1--},
  isbn = {978-1-7138-2954-6}
}

@inproceedings{lakshminarayanan2017simple,
  title={Simple and Scalable Predictive Uncertainty Estimation using Deep Ensembles},
  author={Lakshminarayanan, Balaji and Pritzel, Alexander and Blundell, Charles},
  booktitle={Advances in Neural Information Processing Systems (NeurIPS)},
  volume={30},
  pages = {1--},
  year={2017}
}

@inproceedings{xu2019learning,
  title={Learning an adaptive learning rate schedule},
  author={Xu, Zhen and Dai, Andrew M and Kemp, Jonas and Metz, Luke},
  booktitle={Advances in Neural Information Processing Systems},
  volume={32},
  pages = {1--},
  year={2019}
}

@article{dulac2021challenges,
  title={Challenges of real-world reinforcement learning: definitions, benchmarks and analysis},
  author={Dulac-Arnold, Gabriel and Levine, Nir and Mankowitz, Daniel J and Li, Jerry and Paduraru, Cosmin and Gowal, Sven and Hester, Todd},
  journal={Machine Learning},
  volume={110},
  number={9},
  pages={2419--2468},
  year={2021},
  publisher={Springer},
  doi={10.1007/s10994-021-05961-4}
}

@article{kober2013reinforcement,
  title={Reinforcement learning in robotics: A survey},
  author={Kober, Jens and Bagnell, J Andrew and Peters, Jan},
  journal={The International Journal of Robotics Research},
  volume={32},
  number={11},
  pages={1238--1274},
  year={2013},
  publisher={Sage Publications Sage UK: London, England}
}

\end{document}